# Supporting Literacy Assessment in West Africa: Using State-of-the-Art Speech Models to Evaluate Oral Reading Fluency

**Owen Henkel**[1], **Hannah Horne-Robinson**[1], **Libby Hills**[2], **Bill Roberts**[3], **Joshua McGrane**[4],
Submitted: October 2023

## Abstract

This paper reports on a set of three recent experiments utilizing large-scale speech models to evaluate the oral reading fluency (ORF) of students in Ghana. While ORF is a well-established measure of foundational literacy, assessing it typically requires one-on-one sessions between a student and a trained evaluator, a process that is time-consuming and costly. Automating the evaluation of ORF could support better literacy instruction, particularly in education contexts where formative assessment is uncommon due to large class sizes and limited resources. To our knowledge, this research is among the first to examine the use of the most recent versions of large-scale speech models (Whisper V2 wav2vec2.0 for ORF assessment in the Global South.

We find that Whisper V2 produces transcriptions of Ghanaian students reading aloud with a Word Error Rate of 13.5. This is close to the model's average WER on adult speech (12.8) and would have been considered state-of-the-art for children's speech transcription only a few years ago. We also find that when these transcriptions are used to produce fully automated ORF scores, they closely align with scores generated of expert human graders, with a correlation coefficient of 0.96. Importantly, these results were achieved on a representative dataset (i.e., student with regional accents, recordings taken in actual classrooms), using a free and publicly available speech model out of the box (i.e., no fine-tuning). This suggests that using large-scale speech models to assess ORF may be feasible to implement and scale in lower resource, linguistically diverse educational contexts.

✉   Owen Henkel | owen.henkel@education.ox.ac.uk

[1] University of Oxford, Department of Education
[2] Jacobs Foundation
[2] Legible Labs
[4] University of Melbourne, Graduate School of Education

## Introduction

The rapidly evolving capabilities the newest generation of large language, speech, and multi-modal models (e.g., GPT-4, Bard, Whisper V2, and Wav2Vec 2.0) is an exciting development in the field of Artificial Intelligence and Education. The growth in the size of training datasets and the number of model parameters appears to have dramatically improved both their overall performance benchmark, models' ability to generalize to new dataset, and in some cases has led to the emergence of new and unexpected abilities (Stiennon et al., 2022; Wei et al., 2022). These capabilities suggest that large-scale speech models could be deployed to support a variety of educational activities, such as including real-time translation of a classroom lecture, supporting more collaborative student conversations, or helping with formative assessment of foundational reading skills. However, to date researching the potential use of this new generation of language and speech models has been concentrated in high-income countries, and it is critical to ensure that less well-resourced education systems can also harness the potential advantages of these technologies. We aim to begin to fill this this gap and provide insight into how cutting edge LLMs could play a role in improving literacy assessment in Low and Middle-Income Countries (LMICs).

UNESCO estimates the number of illiterate individuals to be over 800 million, nearly 10% of the global population, the vast majority of whom live in LMICs (Pritchett, 2013). Frequent formative assessment has been shown to be an important component of effectively teaching students to read fluency, as it can inform classroom-level teaching practice, providing real-time feedback to students and teachers (Black & Wiliam, 2009; Nag, 2017). The role of formative assessment is particularly important in Low and Middle-Income Countries (LMICs), where students can develop a mix of component reading skills that would be uncommon in monolingual, well-resourced educational settings (Klaas & Tudell, 2011; Spaull et al., 2020). For instance, a student who has adequate decoding skills but lacks sufficient understanding of the language they are reading might score similarly to a student with weak decoding skills.

One of the most widely used measure of foundational literacy skills is oral reading fluency (ORF). Oral reading fluency - the ability to read a text accurately, with proper expression, at a sufficient rate - is a core component of successfully learning to read (National Reading Panel, 1998; Rasinski et al., 2009). ORF is generally considered a reliable predictor of reading comprehension, particularly in early grades, and it is widely used to assess foundational literacy (Fuchs et al., 2001; Spear-Swerling, 2006). However, assessing ORF typically requires a one-to-one interaction for several minutes, meaning there is significant time, cost, and complexity associated with administering and grading them (Magliano & Graesser, 2012). This means formative literacy conducting these assessments regularly can be challenging for governments and educational institutions, due to large rural populations, unreliable infrastructure, and weak organizational capacity (Dubeck & Gove, 2015). Given the fact that even the largest scale reading assessments are currently only administered infrequently and to a small fraction of students in a given country, automating ORF assessment would be an important development.

In recent years, ASR models have reached human-level performance. However, even state-of-the-art models have failed to perform satisfactorily on child voice datasets (Jain, Barcovschi, Yiwere, Bigioi, et al., 2023). Factors such as a higher pitch, developing vocabulary and irregular pronunciation and acoustic significantly differentiate a child's voice from an adult's voice (Gerosa et al., 2009). In the context of LMICs, there are added complications, most notably limited data representing regional accents (Feng et al., 2024; Nouza et al., 2023).

A variety of approaches have been attempted to enhance the performance of automatic child speech recognition systems, and some contexts-specific carefully finetuned speech model have shown improved performance (Bolaños, Cole, Ward, Tindal, Hasbrouck, et al., 2013; Fan et al., 2022). However, their implementation in real-world educational environments is hindered by two main challenges. Firstly, the need to fine-tune these models restricts their utility and influence. Fine-tuning can be a complicated process that demands advanced technical skills, and the acquisition and labeling of the fine-tuning dataset remains an expensive and labor-intensive undertaking (Jain, Barcovschi, Yiwere, Corcoran, et al., 2023; Radford et al., 2022). Secondly, models that demonstrate "superhuman" performance when trained on a specific dataset can still commit numerous fundamental errors when tested on a different tasks or new dataset, even ones that appear extremely similar to the casual observer (Geirhos et al., 2020). This challenge, typically referred to as domain shift, is often due to the model capitalizing on dataset-specific peculiarities that are not apparent to humans. Hence, for the models to realistically be usable in education systems in much of the global south, they would have to be easy to operate and generalize well across contexts.

Fortunately, addressing the dual challenges of the transfer learning paradigm (i.e., the technical complexity of fine-tuning and failure to generalize to new contexts) has been a major area of research in the past few years. It has been repeatedly found that increasing the scale of language models (e.g., training computation, model parameters, and so forth) can lead to improved performance across numerous downstream tasks (Baevski et al., 2020; Radford et al., 2022; Wei et al., 2022). This fact also appears to apply to speech models, and various researchers have found that conducting supervised pre-training across multiple datasets and domains, leads to models with greater resilience and superior generalization to unseen datasets compared to models trained on a single source (Chan et al., 2021; Narayanan et al., 2018).

In recent work, we explored a related question of whether Large Language Models (GPT-4, Bard, Claude) could evaluate student responses to short-answer reading comprehension questions. We found that GPT-4, with minimal prompt engineering, performed extremely well in evaluating a novel dataset (Quadratic Weighted Kappa 0.923, F1 0.88), described below, substantially outperforming transfer-learning based approaches, and even exceeding expert human raters (Quadratic Weighted Kappa 0.915, F1 0.87). Based on these results, it seemed plausible that large-scale speech models could potentially evaluate ORF, but there has been extremely limited research that tests this hypothesis in Low and Middle-Income Countries.

## Prior Work

### ASR Model architectures and benchmarks

End-to-end automatic speech recognition systems, which utilize a neural network-based architecture that directly convert speech signals into written text, have gained significant attention in recent years (Jain, Barcovschi, Yiwere, Corcoran, et al., 2023; Nouza et al., 2023). The first implementations of this approach typically relied on a fully supervised training paradigm, where models were trained using collections of high-quality audio transcript pairings. However this approach limited the size and diversity of the datasets that were available to train models with (Radford et al., 2022). This resulted in a situation where ASR models matched or exceeded human-level performance on canonical ASR datasets such as LibriSpeech but have struggled with overfitting and domain shift when tested on other datasets (Geirhos et al., 2020; Panayotov et al., 2015).

      In response to these challenges, other researchers proposed unsupervised pre-training methods, most notably wav2vec 2.0 (Baevski et al., 2020). Unsupervised learning approaches have the advantage of being trained on audio without requiring human labels, thus making large datasets of unmarked speech feasible for use. This approach has made it feasible to train models on up to 1,000,000 hours of audio (Zhang et al., 2022). While this approach propelled progress in ASR, they come with a significant drawback. While the pre-trained model learns to encode high-quality representations of speech, because they are purely unsupervised, they lack a decoder with equivalent performance that can map these representations to usable outputs. As a result, a fine-tuning phase is needed for them to perform tasks such as speech recognition. This necessity, unfortunately, curtails their efficacy since fine-tuning can often be a complex process that requires a skilled practitioner (Radford et al., 2022).

      Recognizing the strengths and limitations of both the fully supervised and fully unsupervised approaches, more recent research has started to explore an approach dubbed lightly supervised (Radford et al., 2022). This approach uses both unsupervised and supervised data to train the base model and relaxes the need for all supervised data to be olden-standard human-validated transcripts, allowing the use of audio transcriptions found on the internet such as subtitles of movies and shows, and audio books. Although the trade-off between quality and quantity can often be beneficial, this aspect has been relatively overlooked in speech recognition. The largest and most successful implementation of this approach to date is Whisper V2, which was trained on over 680,000 hours of audio, resulting in a rich collection covering a large range of audio from diverse recording setups, environments, speakers, and languages.

When tested across a variety of ASR datasets Whisper V2 dramatically outperformed wa2Vec2.0, achieving an averaging a WER of 12.8 compared to 29.3.

**ASR in Educational Settings**

Over the past five years, research has made significant strides in ASR for adult speech, and the top performing models now rival human level performance on many tasks, including transcription (Baevski et al., 2020; Radford et al., 2022). However, the models can still struggle when dealing with non-native speakers, regional accents, and children's speech (Feng et al., 2024; Jain, Barcovschi, Yiwere, Corcoran, et al., 2023). First and foremost, children's speech is distinct from adult speech both in terms of their word choice and pronunciation, as well as the acoustic properties due to shorter vocal tracts (Bolaños, Cole, Ward, Tindal, Schwanenflugel, et al., 2013). Due to the limited scale of high-quality child speech datasets available, it has historically been challenging to adapt ASR models to these characteristics. Additionally, background noise in classrooms is ubiquitous and difficult to eliminate from recordings without moving students to a separate room, which would be unfeasible. Consequently, common ASR systems available on the market may underperform when transcribing real-world audio files.

Additionally, differences in accents can also impact the accuracy of transcription, as ASR training datasets may not account for regional differences in accent and pronunciation. For instance, in 2020, researchers found that large-scale commercially available ASR models had a significantly higher WER for black Americans than for white Americans - a fact they attributed to the datasets used to train the models (Koenecke et al., 2020). While there is some evidence that the most recent generation of ASR models have been trained on larger and more diverse datasets, thereby performing better on various cases, they still may not fully account for regional differences in accent and pronunciation. Due to these challenges, implementing ASR in classrooms with young students has been a long-standing challenge in the field of ASR (Gerosa et al., 2009).

Research in further fine-tuning self-supervised speech model learning have shown improvements in child speech recognition. In one set of experiments, Jain et al. (2023) explored using the ASR model wav2vec 2.0 with different pretraining and fine-tuning approaches. Their models outperformed the unmodified wav2vec 2.0 on child speech, when tested on well-established student speech corpora such as MyST Corpus, PFSTAR dataset, CMU Kids dataset. They reported that with as little as 10 hours of child speech data for fine-tuning they were able to achieve a WER below 15, which is on par with adult speech. In a subsequent study, they compared Whisper, a large ASR model released by OpenAI, with fine-tuned self-supervised models like wav2vec 2.0. Whisper, trained with significantly more data and multilingual resources, was expected to improve child speech recognition through fine-tuning. The results showed that fine-tuning Whisper on child speech led to significant improvements in ASR performance compared to the non-fine-tuned Whisper model, reaching parity with human raters. We consider these results (summarized in Figure 1 below) to be state-of-the-art for children's speech transcription. While these and other results appear promising, it remains unclear how

ASR models like Whisper would perform in scenarios which combined children's speech, regional accents, and environmental noise (e.g., background conversations).

**Figure 1**
*Performance of ASR Model on Child-Specific Datasets*

|  | **MyST Corpus** | **PFSTAR dataset** | **CMU Kids dataset** |
| --- | --- | --- | --- |
| **wav2vec 2.0 large** (10 hours finetuning) | 27.17 | 3.50 | 21.35 |
| **wav2vec 2.0 large** (65 hours finetuning) | 7.42 | 2.99 | 14.18 |
| **Whisper V2 Large** (unmodified) | 25.00 | 73.68 | 12.69 |
| **Whisper V2 Large** (10 hours finetuning) | 15.79 | 2.88 | 15.22 |
| **Whisper V2 Large** (65 hours finetuning) | 13.34 | 4.17 | 17.11 |

*Note: Adapted from Jain et. al. 2022*

**Reading Fluency**

Reading fluency refers to the ability to read a text accurately, with proper expression, and at a sufficient rate so that working memory is not overloaded (National Reading Panel, 1998). Non-fluent readers may read words correctly but do so slowly and laboriously, with a stilted and monotonous tone, and struggle with word grouping (Fuchs et al., 2001; Rasinski et al., 2009) Fluent reading is crucial for successful reading because it reflects word knowledge and automaticity, which are both important aspects of proficient reading (Perfetti & Stafura, 2014). As reading becomes more efficient, it requires fewer cognitive resources, allowing for the allocation of resources to tasks such as maintaining the text in working memory, accessing background knowledge, and making inferences (LaBerge & Samuels, 1974).

      Empirical research has demonstrated that without sufficient automaticity, it is impossible to retain a sentence in working memory. For example, if a reader cannot complete a sentence within approximately twelve seconds, they will have forgotten the beginning of the sentence by the end. While there are challenges in measuring silent reading fluency, particularly in classroom settings, much of the research on reading fluency focuses on oral reading fluency (ORF) due to its practicality (Hasbrouck & Tindal, 2006).

      While ORF can be evaluated in various ways, one of the most widely used measures is Words Correct Per Minute (Roehrig et al., 2008). This typically involves asking a student to read

a passage for a set period, while the assessor marks reading errors. When the student finishes, the errors are tallied, subtracted from the total number of words in the original text, and then divided by the time the student spent reading. The resulting score, WCPM, is a combined measure of accuracy and rate, with higher scores considered to be indicative of stronger ORF. These assessments are typically conducted one-on-one with a student and trained evaluators, making it time-consuming and costly to administer. Therefore, there are potential benefits of using NLP and ASR to partially or fully automate ORF assessments (Duong et al., 2011).

## Current Study

In three interrelated experiments we examine the efficacy of publicly available, and free-to-use ASR models to evaluate the oral reading fluency of Ghanaian students. First, we test two top-performing open-source large-scale speech models (Whisper V2 and wav2vec 2.0) on a novel dataset, the *University of Oxford and Rising Academies Reading Comprehension Dataset*. This corpus (subsequently referred to as the Ghana Dataset) consists of a series of literacy evaluation tests conducted on over 150 students from elementary and middle schools in Ghana. Second, in order to better understand the impact of accent and ambient noise on model performance, we compare the accuracy achieved by these models on the Ghana Dataset, to that achieved on dataset of American students. Finally, we use the model transcription to automatically calculate students' ORF scores and compare them with scores generated by excerpted human raters.

### Rising Academies and Oxford University Reading Comprehension Dataset

This dataset is comprised of a series of literacy assessments conducted with students from Rising Academies, a school network in Ghana. The data collected includes the results of various tasks aimed at assessing the reading abilities of 162 students aged between 13 and 18. The reading passages were taken from the 2016 PrePIRLS (now known as PIRLS Literacy), an international comparative study that evaluates the reading skills of fourth-grade students in 60 countries, and which was specifically designed for application in LMICs and nations with lower literacy rates (*Methods and Procedures in PIRLS 2016*, 2018). To evaluate ORF, students were asked to read two short passages aloud, each approximately 100 words long. The oral reading sessions were audio-recorded and later transcribed by human annotators, and then reading errors were marked by expert raters.

**Figure 2**
*Example of ORF Passage and Scoring*

| Rising Academies and Oxford University Reading Comprehension Dataset |
|---|

| Text Being Read | Human Transcript of Audio (Errors marked in red) |
|---|---|
| My name is Kojo.<br>I am seven years old.<br>I have a brother and a sister.<br>My brother's name is Eli<br>and he is five years old.<br>He is learning how to read.<br>My sister's name is Ama<br>and she is the eldest, she is 11.<br>My family lives in Ho,<br>and my father is a farmer.<br>Every Saturday my mother goes to the<br>Market to sell fruits and always takes me,<br>and sister with her. I like going to the market,<br>because when we are not helping mother,<br>my sister and I can explore the market.<br>After market, we go back to home and<br>go out to the farm to visit my father and help him | My name is Kojo.<br>I am seven years old.<br>I have a brother and a sister.<br>My brother's name is eh Eli<br>and his he is five years old.<br>He is learning how to read.<br>My sister's name is Ama<br>and she is the eldest eldest, she is 11.<br>My family lives in Ho,<br>and my my father is a farmer.<br>Every Saturday my mother goes to the<br>market to sell fruits and always takes me,<br>and sister with her. I like going to the market,<br>because when we are when we are not helping mother,<br>my sister and I can expoy the market.<br>After market, we'll go back to home and<br>go out to the farm to visit my father and help him |
| 117 Words in original text | 8 Student reading errors |
| [Audio file of student reading](#) | |

## CU Kids Read and Summarized Story Corpus

The CU Kids Read and Summarized Story Corpus contains speech recordings and transcripts from 106 children summarizing stories they had read. The data was collected in 2003 by the University of Colorado Boulder researchers. The participants were English-speaking elementary students in grades 3–5, approximately 8–12 years old. Each child was asked to read a short story and then provide a spoken retelling summarizing what happened in their own words. No specific guidance was given regarding the length or content of the summaries. The children's spoken summaries were audio-recorded and manually transcribed by expert raters.

**Figure 3**
*Example of ORF Passage and Scoring*

| CU Kids Read and Summarized Story Corpus | |
|---|---|
| Text Being Read | Human Transcript of Audio (Reading errors marked in red) |
| Tim liked to go outside on sunny days.<br>He liked to feel the cool breeze.<br>He liked to hear the wind.<br>He liked to see the white, puffy clouds.<br>Today clouds were racing across the sky.<br>Clouds never stay in one place.<br>Tim wondered where the clouds go.<br>"White puffy clouds come on sunny days,"<br>Tim said.<br>"What else do I know about clouds?"<br>Tim thought about last Monday.<br>Monday was a foggy day.<br>"What is fog?" Tim asked his Mom.<br>"Fog is a layer of clouds near the ground," Mom said.<br>"Where does the fog go?" Tim asked. | Tim liked to go outside on sunny days.<br>He liked to feel the the cool breeze.<br>He liked to hear the wi wind.<br>He liked to see the white, fluffy clouds.<br>Today clouds were racing across the sky.<br>Clouds never stay in one place.<br>Tim wondered where the clouds go.<br>"White fluffy clouds come on sunny days,"<br>Tim said.<br>"What else do I know about clouds?"<br>Tim thought about last Monday.<br>Monday was a foggy day.<br>"What is fog?" Tim asked his Mom.<br>"Fog is a layer of clouds near the ground," Mom said.<br>"Where does the fog go?" Tim asked. |
| 99 Total in original text | 4 Student reading errors |
| [Audio file of student reading](#) | |

**Measuring Model Performance**

Word Error Rate, the standard measure of transcription accuracy, also calculates the relationship between the number of errors and the original length of the text. However, WER only considers the discrepancy between a reference text and a candidate text. In the context of ASR, the WER calculation typically utilizes a high-quality human transcription of the audio as the reference text. This is considered the 'ground-truth', and a model-generated transcript serves as the candidate text. Formally WER is defined as:

$$WER = I + D + S / N$$

where I, D, and S represent the number of word insertions, deletions, and substitutions, identified in the candidate text, and N is the total number of words in the source text (Ali & Renals, 2018). This number is often then multiplied by 100 for ease of interpretation. Hence, WER scores have a definitional minimum score of 0 which indicates that the two documents are identical. While WER can technically exceed 100, the typical range is between 0 and 100, with lower scores being considered better.

    While automatically calculating WCPM removes one potential source of measurement error, raters miscounting errors or miscalculating the final score, it also introduces a new potential source of measurement error, transcription error, which has historically been a major challenge when dealing with audio of children reading in classroom environments.

**Experiment 1: Evaluating Speech Models' Performance on the Ghana Dataset**

We selected 60 audio files from the Ghana Dataset, chosen based on the quality of the audio file and completeness of the story. These audio recordings were meticulously transcribed by an expert human rater and scored manually according to two different ORF scoring methodologies. Subsequently, the audio files were transcribed by two different state-of-the-art ASR models, Whisper V2 large, with zero-shot prompting, and wav2vec 2.0. Importantly, we did not conduct any fine-tuning or modification on these models, using them "out-of-the-box." After generating the model transcripts, we standardized the text of all three transcriptions by removing punctuation and converting to lowercase. We then used JiWER, a simple and fast Python package designed to evaluate automatic speech recognition systems (*JiWER*, n.d.). JiWER calculates the WER of the model transcriptions, relative to the ground truth reference text, which we considered to be the expert human transcription in this case.

**Figure 4**
*WER by model on Ghana Dataset*

|  | **Whisper V2** | **wav2vec 2.0** |
| --- | --- | --- |
| Average WER | 13.5 | 25.2 |
| Range - Max \| Min | 29.2 \| 1.7 | 6.69 \| 5.0 |
| Variance | 0.34 | 0.21 |

Both Whisper V2 and wav2vec 2.0 perform similarly to the average performance on ASR benchmarks reported above. The models' ability to generalize to a novel dataset comprised of students' speaking is notable, due to the historical challenges of transcribing children's speech. However, it is notably lower than the results reported by Jain et al. using bespoke fine-tuned implementations of these models to transcribe children-specific ASR datasets.

Whisper V2 clearly outperforms wav2vec 2.0, which has much higher error rates. However, upon a closer examination of the transcripts two interesting behaviors can be seen. The first is that wav2vec 2.0 preserves reading errors, particularly repetitions, more faithfully (e.g., "he locked, he looked", "had head headed"). Whereas Whisper V2 corrects some repetitions and reading errors. It has been suggested that discrepancies in children's speech production are rectified, or "normalized", more leniently compared to those of non-native adult speakers—transcriptions are typically more direct, including restarts (Feng 2022). The second is that wav2vec 2.0 appears to transcribe some words phonetically, relative to an American accent. For instance, the way the Ghanaian student says the word "rushing" is phonetically like how an average American might say the word "Russian". Other examples include "rog", for "rock" and the space between "river" and "bed".

These differences are likely attributable to distinct training approaches. wav2vec 2.0 purely unsupervised training approach, makes it more sensitive to the acoustic properties of the reading. In contrast, Whisper V2's lightly supervised approach means that the model is trained on transcript-audio pairs, often of mixed quality, which often omit repeated words, and capture what speaker was presumed to be trying to say(Radford et al., 2022). As a result, Whisper has likely learned this from its training data, and the transcripts may edit out repetitions or verbal stumbles. Artifacts of this can be seen by comparing transcripts, with differences highlighted in yellow.

**Figure 5**

*Alternate Transcripts of Same Audio Clip*

| Whisper V2 large | wav2vec 2.0 (transcription differences highlighted in yellow) |
|---|---|
| a farmer went out one day to search for a lost scarf he went to the valley and searched by the riverbed among the reeds behind the rocks and in the rushing water he climbed the slopes of the high mountain with its rocky cliffs he looked behind a large rock in a case the calf had headed there to escape the storm and that | a farmer went out one day to search for a lost scarf he went to the valley and serched by the river bed among the reeds behind the rocks and in the russian water he climbed the slopes of the high mountain with its rocky cliffs he locked he looked behind a large rog in a case the calf had head headed there to escape the storm and that |
| [Audio file of student reading](#) | |

## Experiment 2: Evaluating Speech Models' Performance on CU Dataset

The recordings in the Ghana Dataset differ from many other children specific ASR datasets in that the recordings were made classrooms, the vast majority of students spoke English a second (or third) language, and because the local accent is distinct from many North American accents. To better understand how transcription accuracy was affected by these factors, we also evaluated Whisper V2 on transcribed recordings from the CU Dataset. To do so, we selected 45 audio recordings, representing a mix of age and ability levels, which already had high-quality human generated transcripts. We calculated the WER of the model generated transcript relative to the human transcript, which we considered the ground truth.

**Figure 6**

*Whisper V2 WER by Dataset*

|  | Ghana Dataset | CU Dataset |
|---|---|---|
| **Average WER** | 13.5 | 10.2 |

| Range - Max \| Min | 29.2 \| 1.7 | 31.0 \| 0.0 |
|---|---|---|
| Variance | 0.34 | 0.75 |

The model exhibited a lower WER on the CU Dataset, compared to the Ghana Dataset, which is consistent with prior research. That ASR systems can have difficulties adapting to different speech due to variables such as gender, age, speech disorders, ethnicity, and accents. If the model was originally trained on speech derived from native speakers of a "standard" variant of a specific language, the model might unintentionally be biased against the speech of non-native speakers as well as speakers from different regional or sociolinguistic variations of the language (Feng et al., 2024; Nouza et al., 2023). However, unlike previous studies that reported large gaps in WER - (Koenecke et al., 2020) reported a gap in WER of over 20 gap between black and white speakers - in this case the gap of 3.3 is modest. Additionally, a portion of this 3.3 WER gap may be partially attributable to the additional background noise in the Ghana dataset. The relatively small decrease in accuracy may be due to the larger, more varied, multilingual data sets the newest generation of speech models are trained on. However, given the importance of issues of bias and equity in speech models, this is an important area for continued research.

## Experiment 3: Automatically Calculating WCPM From a Transcript

The final experiment examines how closely a fully automated Words Correct Per Minute scores matches scores generated by expert human raters. As discussed above WCPM is widely used measure of ORF and combines both reading accuracy and reading rate.

To generate human WCPM scores the raters (MSc student from the Department of Education) carefully listened to the audio clips while reading along with the source story and counted student reading errors according to the standard WER protocol (insertions, deletions, and substitutions). After double checking their count of errors, the number of errors was subtracted from the total story length, and was then divided by the duration of the audio clip in seconds, and multiplied by 60.

*WCPM = (total words in story - total errors) x (60 / duration of audio clip in seconds)*

For the fully automated estimate of WCPM, Whisper V2 was used to transcribe the audio clips. Then the JiWER Python package to calculate the WER between the original story text and the transcript. The WER score, was subtracted from 1, and multiplied by the total number of words in the story to arrive at the total number of correctly read words. The number was divided by the duration of the audio clip in seconds and multiplied by 60

*WCPM = (total words in story* (1-WER)) x (60 / duration of audio clip in seconds)*

The overall average WCPM estimates were quite close when using the expert human and the fully automated approach, 113 compared to 110. This difference would be considered small when comparing two separate groups of students, and from a statistical perspective, the difference would not be statistically significant (i.e., the difference could likely be due to normal variation). We also we also considered other measures to evaluate whether there was high variability between the two approaches at the student level, which is how WCPM measures should be used.

To do so first calculated the correlation coefficient between the two sets WCPM scores finding an extremely high value of 0.964. Next, we calculated the absolute value of the difference between the fully human and fully automated method for each student. This measure can give an approximate sense of how far "off" the fully automated approach is from the true value for a given student. The average difference of 5 WCPM would again be considered small in the context of WCPM measures.

**Figure 7**
*Comparing WCPM Scores by Method*

|  | **Expert Human** | **Fully Automated** |
|---|---|---|
| Average | 113 | 110 |
| Correlation Coefficient | 0.964 ||
| Average Change in WCPM | 5 ||

**Limitations**

While this study provides valuable insights into the potential of ASR models for the automated evaluation of ORF, it is not without its limitations. Firstly, the study was conducted using only two specific ASR models, Whisper V2 and wav2vec 2.0. While these models are considered state-of-the-art, the findings may not be generalizable to other ASR models. Secondly, the study used a relatively small sample size from two specific datasets, one from Ghana and one from the United States. Thirdly, the study found evidence of model bias against underrepresented accents, which could limit the effectiveness of these models in diverse educational settings. A particularly promising area might be the further fine-tuning of ASR models to make them more sensitive to a) regional accents and b) the need to preserve reading errors, rather than correct them, as it appears that Whisper V2 may do.

Finally, these experiments only evaluate the potential for automating ORF in English. Historically, there has been a noticeable gap in model performance based on language and improving ASR for underserved languages is an active area of research. Despite improvements, a larger scale model was not evaluated. Additionally, there are active debates about the accuracy of ORF measures across different languages, even when conducted by expert human assessors. In short, an important area for further research would be the replication of these findings using larger data set.

## Conclusion

The primary objective of this study was to explore the feasibility and efficacy of employing ASR models for the automated evaluation of Oral Reading Fluency. The study was particularly focused on addressing two central challenges associated with the automation of ORF measurement in LMICs: the quality of transcription and the conversion of transcription into WCPM. The experimental design of the study involved the utilization of two state-of-the-art ASR models, namely Whisper V2 and wav2vec 2.0, to transcribe audio files of students' oral reading from two distinct datasets: a novel dataset from Ghana and the CU Datasets and Summarized Story Corpus from the United States. The transcriptions generated by these models were subsequently used to calculate a WCPM score for each student, which was then compared to the score derived from human transcripts.

The key findings of the research are multifaceted and provide valuable insights into the potential of ASR models in the context of ORF assessment. Firstly, the Whisper V2 model demonstrated superior performance in transcribing the audio files, exhibiting a lower WER compared to the wav2vec 2.0 model. However, the wav2vec 2.0 model was found to preserve reading errors more faithfully and transcribe certain words phonetically, which could be attributed to its distinct training approach. Secondly, the research revealed that the WER of the model transcriptions was marginally higher for the Ghana Dataset compared to the CU Dataset, suggesting a degree of model bias against underrepresented accents. Lastly, the study found that the fully automated approach to calculating WCPM from the transcripts yielded results that were remarkably close to the expert human calculations, with a high correlation coefficient and a small average difference in WCPM.

The implications of these findings for the automatic assessment of ORF are substantial. The results suggest that ASR models have the potential to accurately transcribe student oral reading and calculate WCPM scores, thereby enhancing the efficiency of ORF assessment and reducing the reliance on human evaluators. This could be particularly beneficial in LMICs, where the systematic and regular assessment of literacy skills can be challenging due to factors such as large rural populations, unreliable infrastructure, and limited organizational capacity.
In the broader context of formative assessment, the findings of this research suggest that ASR models could be employed to assist in the evaluation of student work across a variety of contexts. The demonstrated ability of these models to accurately transcribe student speech and calculate performance metrics could potentially be applied to other areas of assessment, such as oral presentations or language proficiency tests. However, the research also underscores the need for further investigation to address the challenges associated with model bias against underrepresented accents and the technical complexity of fine-tuning these models. The findings of this study, therefore, not only contribute to the existing body of knowledge on the application of ASR models in educational settings but also highlight potential avenues for future research in this area.

**Declarations**

The authors have no relevant financial or non-financial interests to disclose.